\PassOptionsToPackage{dvipsnames,table}{xcolor}
\documentclass[10pt]{article} 

\usepackage[accepted]{rlj} 

\usepackage{amssymb}            
\usepackage{mathtools}          
\usepackage{mathrsfs}           
\usepackage{graphicx}           
\usepackage{subcaption}         
\usepackage[space]{grffile}     
\usepackage{url}                
\usepackage{lipsum}             

\usepackage{cleveref}
\usepackage{amsthm}
\usepackage{booktabs}
\usepackage{multicol}
\usepackage{multirow}
\usepackage[export]{adjustbox}
\usepackage[ruled,vlined]{algorithm2e}
\usepackage{scalefnt}

\newcommand{\bs}{\mathbf{s}}
\newcommand{\ba}{\mathbf{a}}

\newcommand{\mcS}{\mathcal{S}}
\newcommand{\mcA}{\mathcal{A}}
\newcommand{\mcP}{\mathcal{P}}
\newcommand{\R}{\mathbb{R}}

\newcommand{\dqs}{\textsc{\scalefont{1.2}dqs}\xspace}
\DeclareMathOperator*{\argmax}{argmax}

\definecolor{highlight}{RGB}{225,230,255}
\newcommand{\highlight}[1]{\cellcolor{blue!10}{#1}}

\title{Sampling from Energy-based Policies using Diffusion}

\setrunningtitle{Sampling from Energy-based Policies using Diffusion}

\author{Vineet Jain, Tara Akhound-Sadegh, Siamak Ravanbakhsh}

\emails{\{jain.vineet, tara.akhoundsadegh, siamak.ravanbakhsh\}@mila.quebec}

\affiliations{
\textbf{School of Computer Science, McGill University}\\
\textbf{Mila - Quebec AI Institute}
}

\contribution{
    We develop a novel actor-critic reinforcement learning algorithm such that the policy samples actions from the Boltzmann distribution of the Q-function. We achieve this by using a diffusion model to parameterize the policy that explicitly learns the score function of the target Boltzmann density.
    }
    {
    Boltzmann policies are a popular choice in discrete action spaces. However, sampling from these policies in continuous action spaces is generally intractable. Prior work \citep{psenka2023learning} used Langevin sampling to address this challenge. Other applications of diffusion models \citep{wang2024diffusion} backpropagate the gradient through the entire diffusion chain to maximize Q-values. To the best of our knowledge, our method is the first to use diffusion models to explicitly sample from Boltzmann policies.
    }

\contribution{
    Experiments on continuous control tasks demonstrate improved sample efficiency of our method compared to relevant baselines.
    }
    {
    We observe higher returns with fewer number of environment interactions (compared to our baselines) on a majority of tasks.
    }

\contribution{
    We demonstrate that our proposed method can learn multimodal behaviors in maze navigation tasks. 
    }
    {
    Our setup consists of a maze with two possible goals. Multimodality in this context refers to the ability of an agent to reach both goals from some initial state, and discover multiple paths (if they exist) to a goal. We qualitatively examine the trajectories of a trained agent and compare them with respect to goal coverage and diversity of paths.
    }


\summary{Energy-based policies offer a flexible framework for modeling complex, multimodal behaviors in reinforcement learning (RL). In maximum entropy RL, the optimal policy is a Boltzmann distribution derived from the soft Q-function, but direct sampling from this distribution in continuous action spaces is computationally intractable. As a result, existing methods typically use simpler parametric distributions, like Gaussians, for policy representation — limiting their ability to capture the full complexity of multimodal action distributions. In this paper, we introduce a diffusion-based approach for sampling from energy-based policies, where the negative Q-function defines the energy function. Based on this approach, we propose an actor-critic method called Diffusion Q-Sampling (\dqs) that enables more expressive policy representations, allowing stable learning in diverse environments. We show that our approach enhances sample efficiency in continuous control tasks and captures multimodal behaviors, addressing key limitations of existing methods.
}

\begin{document}

\maketitle  

\begin{abstract}
Energy-based policies offer a flexible framework for modeling complex, multimodal behaviors in reinforcement learning (RL). In maximum entropy RL, the optimal policy is a Boltzmann distribution derived from the soft Q-function, but direct sampling from this distribution in continuous action spaces is computationally intractable. As a result, existing methods typically use simpler parametric distributions, like Gaussians, for policy representation --- limiting their ability to capture the full complexity of multimodal action distributions. In this paper, we introduce a diffusion-based approach for sampling from energy-based policies, where the negative Q-function defines the energy function. Based on this approach, we propose an actor-critic method called \textsc{Diffusion Q-Sampling} (\dqs) that enables more expressive policy representations, allowing stable learning in diverse environments. We show that our approach enhances sample efficiency in continuous control tasks and captures multimodal behaviors, addressing key limitations of existing methods. Code is available at \href{https://github.com/vineetjain96/Diffusion_Q_Sampling.git}{https://github.com/vineetjain96/Diffusion\_Q\_Sampling.git}
\end{abstract}


\section{Introduction}

Deep reinforcement learning (RL) is a powerful paradigm for learning complex behaviors in diverse domains, from strategy-oriented games \citep{silver2016mastering, berner2019dota, schrittwieser2020mastering} to fine-grained control in robotics \citep{kober2013reinforcement, sunderhauf2018limits, wu2023daydreamer}. In the RL framework, an agent learns to make decisions by interacting with an environment and receiving feedback in the form of reward. The agent aims to learn a policy that maximizes the cumulative sum of rewards over time by exploring actions and exploiting known information about the environment's dynamics.

The parameterization of the policy is a crucial design choice for any RL algorithm. Under the conventional notion of optimality, under full observability, there always exists an optimal deterministic policy that maximizes the long-term return \citep{sutton2018reinforcement}. 
However, this is only true when the agent has explored sufficiently and has nothing to learn about the environment. Exploration requires a stochastic policy to experiment with different potentially rewarding actions.
Moreover, even in the exploitation phase, there may be more than one way of performing a task, and we might be interested in mastering all of them. This diversification is motivated by the robustness of the resulting policy to environment changes; if certain pathways for achieving a task become infeasible due to a change of the dynamics or reward, some may remain feasible, and the agent has an easier time in adapting to this change by exploiting and improving the viable options.
This argument also suggests that such policies can serve as effective initialization for fine-tuning on specific tasks.

While exploration, diversity and robustness motivate stochastic policies, representing such policies in continuous action spaces remains challenging.
As a result, stochasticity is often introduced by noise injection \citep{Lillicrap2015ContinuousCW} or using an arbitrary parametric family \citep{Schulman2015TrustRP} which lacks expressivity. Orthogonal to the difficulty of representing such policies is their training objective; policies are often optimized to maximize the Q-function, and stochasticity is introduced to encourage exploration as an afterthought. However, our argument for stochasticity favours multi-modal policies; instead of learning the \textit{single best} way to solve a task, we want to learn \textit{all reasonably good} ways to solve the task.

We address both of these issues by explicitly sampling from energy-based policies of the form, 
\begin{equation*}
    \pi(a \mid s) \propto \exp(Q^\pi(s,a)).
\end{equation*}
This is also known as the Boltzmann distribution of the Q-function. The optimal policy in the maximum entropy RL framework is also known to be of this form, except it uses the soft Q-function \citep{haarnoja2017reinforcement}. 
Such a policy has several benefits. First, it offers a principled way to balance exploration and exploitation in continuous action spaces. By sampling from this distribution, the policy still prioritizes actions with high Q-values but also has a non-zero probability of sampling sub-optimal actions. While the use of Boltzmann policies is common in the discrete setting, it is challenging in continuous spaces. This sampling problem is often tackled with Markov Chain Monte Carlo (MCMC) techniques, which can be computationally expensive and suffer from exponential mixing time. Second, this formulation naturally incorporates multimodal behavior, since the policy can sample one of multiple viable actions at any given state. However, such policies are generally intractable to sample from in continuous action spaces, requiring approximations in policy parameterization often at the cost of expressivity.

\begin{figure}[!t]
    \centering
    \includegraphics[width=\linewidth]{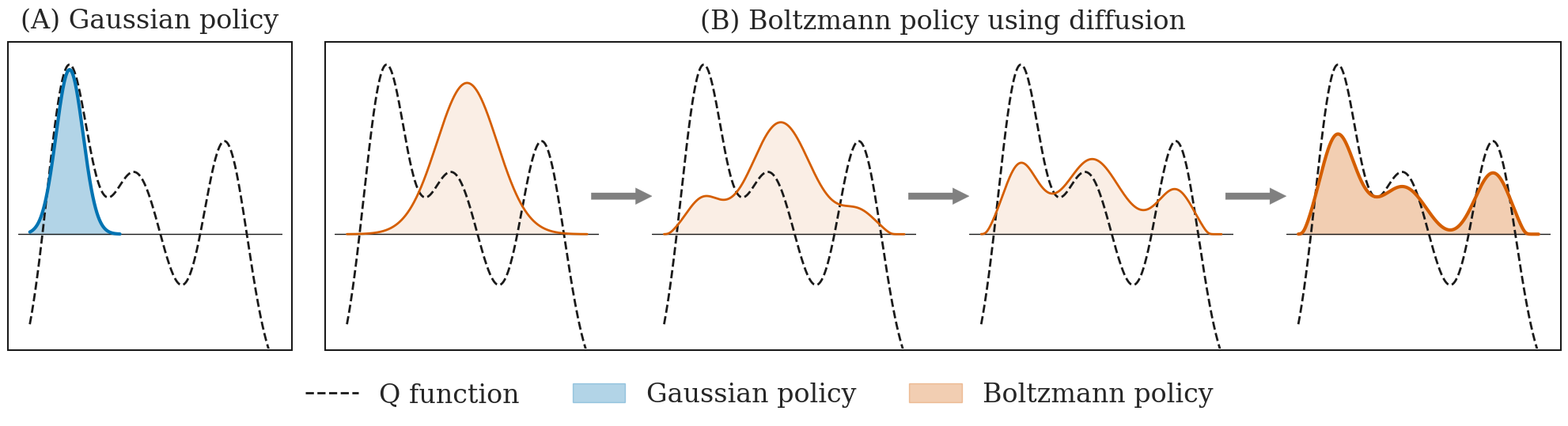}
    \caption{(A) Gaussian policies can only cover a single mode of the action-value landscape, leaving other high-value actions unexplored. (B) Our method trains a diffusion policy that iteratively denoises samples from a standard Normal distribution into samples from the Boltzmann density $ \propto \exp(Q^\pi(s,a))$. The resulting policy can express rich, multimodal action distributions.}
    \vspace{-0.5em}
    \label{fig:policy}
\end{figure}

Diffusion models offer a potential solution to the policy parameterization problem since they are expressive and can produce high-quality samples from complex distributions. Indeed, they have been extensively applied to solve sequential decision-making tasks, especially in offline settings where they can model multimodal datasets from suboptimal policies or diverse human demonstrations. 
A few studies have applied these models in the online setting, focusing on deriving training objectives for policy optimization via diffusion. \citet{yang2023policy} uses the gradient of the Q-function to refine actions sampled from a diffusion policy; however, the exact form of the policy is unspecified, and it is unknown what distribution the diffusion models sample from. \citet{psenka2023learning} samples from the Boltzmann distribution of the Q-function using Langevin dynamics, which may suffer from insufficient mode coverage in high dimensions. \citet{wang2024diffusion} uses diffusion policy within the maximum entropy framework, where the entropy is approximated using a mixture of Gaussians. In contrast, our approach directly samples from $\exp(Q^\pi(s,a))$ by constructing a diffusion process that estimates the score of this target Boltzmann density at different noise scales.

Our contributions in this work are as follows:
\begin{itemize}[leftmargin=15pt,topsep=-0.5ex,itemsep=-0.2ex,partopsep=0ex,parsep=1ex]
    \item We propose a novel actor-critic algorithm, \textsc{Diffusion Q-Sampling} (\dqs), for sequential decision-making using diffusion models for sampling from energy-based policies.
    \item We show that \dqs approximately samples from the Boltzmann density of the Q-function. 
    \item We demonstrate that \dqs is more sample efficient compared to both classical actor-critic methods and more recent diffusion-based methods in continuous control tasks.
    \item We demonstrate that \dqs can learn multimodal behaviors in maze navigation tasks.
\end{itemize}

\section{Related work}
\label{sec:related}

Our work is related to two distinct sub-areas of reinforcement learning: the relatively new and actively explored line of work on applying diffusion models in the RL setting, and the classical maximum entropy RL framework.

\paragraph{Diffusion models in RL. } Early work on applying diffusion models for RL was focused on behavior cloning in the offline setting \citep{chi2023diffusion}. This setting more closely matches the original purpose of diffusion models - to match a distribution to a given dataset. \citet{janner2022planning, ajay2023conditional} use a diffusion model trained on the offline data for trajectory optimization, while \citet{reuss2023goal, jainlearning} apply diffusion models to offline goal-reaching tasks by additionally conditioning the score function on the goal. Within the behavior cloning setting, there is some existing work on learning a stochastic state dynamics model using diffusion \citep{li2022exponential}.

Beyond behavior cloning, offline RL methods incorporate elements from Q-learning to learn a value function from the offline dataset and leverage it to improve the diffusion policy. Most works in this sub-area parameterize the policy using a diffusion model and propose different training objectives to train the diffusion policy. \citet{wang2023diffusion, kang2024efficient} add a Q-function maximizing term to the diffusion training objective, and \citet{hansen2023idql} use an actor-critic framework based on a diffusion policy and Implicit Q-learning \citep{kostrikov2021offline}. \citet{lu2023contrastive} take an energy-guidance approach, where they frame the problem as using the Q-function to guide the behavior cloning policy to high reward trajectories.

The application of diffusion models has been relatively less explored in the online setting \citep{dingconsistency}. DIPO \citep{yang2023policy} modifies actions in the replay buffer based on the gradient of the Q-function, which are then used to train a diffusion model. QSM \citep{psenka2023learning} uses Langevin sampling with a neural network trained to match the gradient of the Q-function.

\paragraph{Maximum entropy RL. } In contrast to standard RL, where the goal is to maximize expected returns, in the maximum entropy RL framework, the value function is augmented by Shannon entropy of the policy. \citet{ziebart2008maximum} applied such an approach in the context of inverse reinforcement learning and \citet{haarnoja2017reinforcement} generalized this approach by presenting soft Q-learning to learn energy-based policies. A follow-up work, \citet{haarnoja2018softa}, presented the well-known soft actor-critic (SAC) algorithm. This line of work proposes to learn a soft value function by adding the entropy of the policy to the reward. The optimal policy within this framework is a Boltzmann distribution, where actions are sampled based on the exponentiated soft Q-values. Some recent works use diffusion models within this framework, such as DACER \citep{wang2024diffusion}, which uses a diffusion policy to represent a maximum entropy policy and estimates the entropy using a mixture of Gaussians. A recent work \citep{ishfaqlangevin} uses Langevin Monte Carlo to improve critic learning through uncertainty estimation over policy optimization.

A separate but related line of work on generative flow networks (GFlowNets), originally defined in the discrete case \citep{bengio2021flow, bengio2023gflownet},  learns a policy that samples terminal states proportional to the Boltzmann density corresponding to some energy function. They have been extended to the continuous setting \citep{lahlou2023theory} and under certain assumptions, they are equivalent to maximum entropy RL \citep{tiapkin2024generative, deleu2024discrete}. They can effectively sample from the target distribution using off-policy exploration, however, they encounter challenges in credit assignment and exploration efficiency \citep{malkin2022trajectory, madan2023learning, rector2023thompson, shen2023towards}. Our approach is distinct as we sample the action at each step from the Boltzmann density of the Q-function, instead of the terminal states based on the reward.



\section{Preliminaries}

\subsection{Reinforcement learning}
We consider a finite-horizon Markov Decision Process (MDP) denoted by the tuple $(\mcS, \mcA, \mcP, r, \gamma)$, where the state space $\mcS$ and action space $\mcA$ are continuous. $\mcP : \mcS \times \mcA \times \mcS \rightarrow [0,1]$ denotes the transition probability of the next state $\bs_{t+1} \in \mcS$ given the current state $\bs_t \in \mcS$ and action $\ba_t \in \mcA$. The reward function $r : \mcS \times \mcA \rightarrow \R$ is assumed to be bounded $r(\bs,\ba) \in [r_{\mathrm{min}}, r_\mathrm{max}]$. $\gamma \in [0,1]$ is the discount factor.

A policy $\pi : \mcS \rightarrow \mcA$ produces an action for every state $s \in \mcS$. In the standard RL framework, the objective is to learn a policy that maximizes the expected sum of rewards $\sum_t \mathbb{E}_{a_t \sim \pi(\cdot \mid s_t), s_{t+1} \sim \mcP(\cdot \mid s_t, a_t)}\left[\gamma^t r(s_t, a_t)\right]$.

The actor-critic framework is commonly used for learning such policies. It involves optimizing a policy (the actor) to choose actions that maximize the action value function, also known as the Q-function (the critic). The Q-function is defined as the sum of expected future rewards starting from a given state-action pair, and thereafter following some policy $\pi$ until terminal time step $T$:
\begin{equation*}
    Q^\pi(s,a) = \mathbb{E}_\pi\left[\sum_{k=t}^{T} \gamma^{k-t} r(s_k,a_k) \mid s_t = s, a_t = a\right].
\end{equation*}
The optimal policy is defined as the policy that maximizes the sum of rewards along a trajectory:
\begin{equation*}
    \pi^* = \argmax_\pi \mathbb{E}_\pi\left[\sum_{t=0}^T \gamma^t r(s_t,a_t)\right].
\end{equation*}

\subsection{Diffusion models}

\paragraph{Denoising diffusion.} Denoising diffusion \citep{dinh2016density, ho2020denoising, song2021score} refers to a class of generative models which relies on a stochastic process which progressively transforms the target data distribution to a Gaussian distribution. The time-reversal of this diffusion process gives the generative process which can be used to transform noise into samples from the target data distribution. 

The forward noising process is a stochastic differential equation:
\begin{equation}
\label{eq:noising_sde}
    dx_\tau = -\alpha(\tau ) x_\tau  d\tau  + g(\tau ) dw_\tau\,,
\end{equation}
where $w_\tau $ denotes Brownian motion. 
In this paper, we consider the Variance Exploding (VE) SDE where the decay rate, $\alpha$, is set to $\alpha(\tau )=0$. This noising process starts with samples from the target density $x_0 \sim p_0$ and progressively adds noise to them over a diffusion time interval $\tau \in [0, 1]$. The marginal probability distribution at time $\tau $ is denoted by $p_\tau$ and is the convolution of the target density $p_0$ with a normal distribution with a time-dependent variance, $\sigma_\tau ^2$. For the VE setting we consider, these marginal distributions are given by:
\begin{equation}
    p_\tau(x_\tau) = \int_0^\tau p_0(x_0) \mathcal{N}(x_\tau; x_0, \sigma_\tau^2) dx_0\,,
\end{equation}
where the variance is related to the diffusion coefficient, $g(\tau )$ via $\sigma_\tau ^2 = \int_0^\tau  g(\xi)^2 d\xi$.

The generative process corresponding to the corresponding to \Cref{eq:noising_sde} is an SDE with Brownian motion $\bar{w}_\tau$, given by:
\begin{equation}
    dx_\tau = [-\alpha(\tau ) x_\tau - g(\tau)^2 \nabla \log p_\tau(x_\tau)] d\tau  + g(\tau ) d\bar{w}_\tau \,.
\label{eq:reverse}
\end{equation}
Therefore, to be able to generate data, we need to estimate the score of the intermediate distributions, $\nabla \log p_\tau(x_\tau)$. 

\paragraph{Iterated Denoising Energy Matching.} Recently, \citet{akhoundsadegh2024iterated} proposed an algorithm known as iDEM (Iterated Denoising Energy Matching) for sampling from a Boltzmann-type target distribution, $p_0(x) \propto \exp(-\mathcal{E}(x))$, where $\mathcal{E}$ denotes the energy. iDEM is a diffusion-based neural sampler, which estimates the diffusion score, $\nabla \log p_\tau$ using a Monte Carlo estimator.
Given the VE diffusion path defined above, iDEM rewrites the score of the marginal densities as:
\begin{equation}
    \begin{split}
        \nabla \log p_\tau &= \frac{\int \nabla \exp(-\mathcal{E}(x_0)) \mathcal{N}(x_\tau; x_0, \sigma_\tau^2) dx_0}{\int \exp(-\mathcal{E}(x_0)) \mathcal{N}(x_\tau; x_0, \sigma_\tau^2) dx_0} \\
        &= \frac{\mathbb{E}_{\tilde{x} \sim \mathcal{N}(x_\tau, \sigma_\tau^2)} \big[ \nabla \exp(-\mathcal{E}(x) \big]}{\mathbb{E}_{\tilde{x} \sim \mathcal{N}(x_\tau, \sigma_\tau^2}) \big[ \exp(-\mathcal{E}(x) \big]}.
    \end{split}
\end{equation}

By observing that the above equation can be written as the gradient of a logarithm leads to the $K$-sample Monte-Carlo estimator of the score:
\begin{equation}
\label{eq:idem}
    S_k(x_\tau, \tau) = \nabla_{x_{\tau}} \log \sum_{i=1}^K \exp \big(-\mathcal{E}(\tilde{x}^{(i)}) \big), \quad \tilde{x}^{(i)} \sim \mathcal{N}(x_\tau, \sigma_\tau^2).
\end{equation}

A score-network, $f_\phi$ is trained to regress to the MC estimator, $S_K$. The network is trained using a bi-level iterative scheme: (1) in the outer-loop a replay buffer is populated with samples that are generated using the model and (2) in the inner-loop the network is regressed $S_k(x_\tau, \tau)$ where $x_\tau$  are noised samples from the replay buffer.

\section{An actor-critic algorithm for Boltzmann policies}

Our objective is to learn general policies of the form
    $\pi(\ba \mid \bs) \propto  \exp(-\mathcal{E}(\bs, \ba))$,
where $\mathcal{E}$ represents an energy function which specifies the desirability of state-action pairs. By setting the Q-function, $Q(\bs,\ba)$ as the negative energy, we get what is known as the Boltzmann policy:
\begin{equation}
    \pi(\ba \mid \bs; T) = \frac{\exp(\frac{1}{T}Q(\bs,\ba))}{\int_\ba \exp(\frac{1}{T}Q(\bs, \ba)) \mathrm{d}\ba}.
\label{eq:boltzmann}
\end{equation}
Choosing such a policy gives us a principled way to balance exploration and exploitation. Specifically, by scaling the energy function with a temperature parameter $T$ and annealing it to zero, we get a policy that initially explores to collect more information about the environment and over time exploits the knowledge it has gained.

\subsection{Diffusion Q-Sampling}

We propose an off-policy actor-critic algorithm, which we call \textsc{Diffusion Q-Sampling} (\dqs), based on the above formulation. Being an off-policy method means \dqs can reuse past interactions with the environment by storing them in a replay buffer $\mathcal{D}$, improving sample efficiency.

Let $Q_\theta$ denote the Q-function and $\pi_\phi$ a parametric policy, where $\theta, \phi$ represent the parameters of a neural network.
The Q-function is learned using standard temporal difference learning:
\begin{align}
    J(\theta) = \mathbb{E}_{(\bs_t,\ba_t) \sim \mathcal{D}}\left[\left(Q_\theta(\bs_t,\ba_t) - \hat{Q}(\bs_t,\ba_t)\right)^2\right],
\label{eq:q_loss}
\end{align}
where
\begin{align*}
    \hat{Q}(\bs_t,\ba_t) = r(\bs_t,\ba_t) + \gamma\,\mathbb{E}_{\bs_{t+1} \sim \mcP, \ba_{t+1} \sim \pi_\phi} \left[Q_{\bar{\theta}} (\bs_{t+1}, \ba_{t+1})\right].
\end{align*}

The target Q-values, $\hat{Q}$, make use of a target Q-network denoted by $Q_{\bar{\theta}}$, where the parameters $\bar{\theta}$ are usually an exponentially moving average of the Q-network parameters $\theta$. Also, in practice, the expectation over next states $\bs_{t+1}$ is estimated using only a single sample.

We parameterize the policy using a diffusion process and use iDEM \citep{akhoundsadegh2024iterated} to sample actions from the target density $\pi(\cdot | \bs_t) \propto \exp(Q_\theta(\bs_t, \ba_t))$.

\begin{algorithm}[t]
\caption{Diffusion Q-Sampling (\dqs)}
\label{alg:dqs}
\DontPrintSemicolon
\SetAlgoLined
\SetKwInput{KwInit}{Initialize}
\KwInit{Initialize Q-function parameters $\theta$, policy parameters $\phi$, target network $\bar{\theta} \leftarrow \theta$, replay buffer $\mathcal{D}$}
\For{each iteration}{
    \tcp{Environment Interaction}
    \For{each environment step}{
        Observe state $\bs_t$ and sample action $\ba_t$ via reverse diffusion using $f_\phi$\;
        Execute $\ba_t$, observe reward $r_t$ and next state $\bs_{t+1}$\;
        Store transition $(\bs_t, \ba_t, r_t, \bs_{t+1})$ in $\mathcal{D}$\;
    }
    \tcp{Parameter Updates}
    \For{each gradient step}{
        Sample minibatch $\mathcal{B} = \{(\bs_t, \ba_t, r_t, \bs_{t+1})\}$ from $\mathcal{D}$\;
        \tcp{Update Q-function parameters $\theta$}
        Compute target Q-values:
        $\hat{Q}_t = r_t + \gamma Q_{\bar{\theta}}(\bs_{t+1}, \ba_{t+1}),\quad \ba_{t+1} \sim \pi_\phi(\bs_{t+1})$\;
        Update $\theta$ by minimizing:
        $J(\theta) = \frac{1}{|\mathcal{B}|} \sum_{\mathcal{B}} \left(Q_\theta(\bs_t, \ba_t) - \hat{Q}_t\right)^2$\;
        \tcp{Update policy parameters $\phi$}
        \For{each $(\bs_t, \ba_t)$ in $\mathcal{B}$}{
            Sample diffusion time $\tau \sim \mathcal{U}[0, 1]$\;
            Sample noisy action $\ba_{t,\tau} \sim \mathcal{N}(\ba_t, \sigma^2_\tau \mathbf{I})$\;
            Sample $\{\tilde{\ba}_t^{(i)}\}_{i=1}^K$, where $\tilde{\ba}_t^{(i)} \sim \mathcal{N}(\ba_{t,\tau}, \sigma^2_\tau \mathbf{I})$\;
            Estimate score:
            $S_t = \nabla_{\ba_{t,\tau}} \log \sum_{i=1}^K \exp\left(Q_\theta(\bs_t, \tilde{\ba}_t^{(i)})\right)$\;
            Update $\phi$ by minimizing:
            $J(\phi) = \left\| f_\phi(\bs_t, \ba_{t,\tau}, \tau) - S_t \right\|^2$\;
        }
        \tcp{Update target network}
        Update $\bar{\theta} \leftarrow \eta\, \theta + (1 - \eta)\, \bar{\theta}$\;
    }
}
\end{algorithm}

\begin{figure}[b!]
    \centering
    \includegraphics[width=0.85\textwidth]{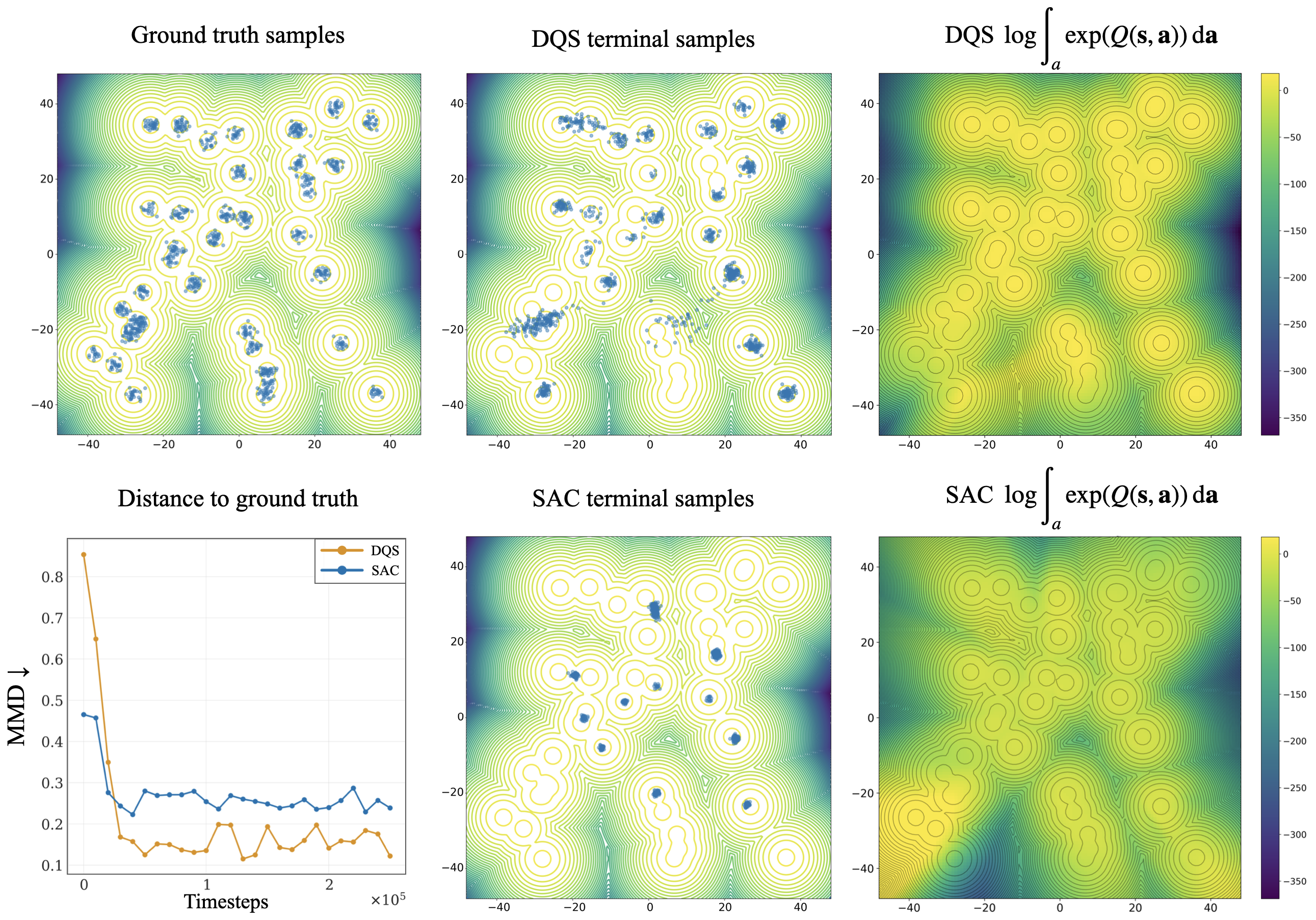}
    \caption{Ground truth, \dqs, and SAC terminal samples for the Gaussian mixture experiment. The right panels show the log partition function based on the learned Q-functions for \dqs and SAC, with contours showing the ground truth density for reference. We evaluate the samples using maximum mean discrepancy (MMD) with the ground truth samples (bottom left).}
    \label{fig:gmm}
\end{figure}

\paragraph{Forward process.} Given $(\bs_t, \ba_t) \in \mcS \times \mcA$, we progressively add Gaussian noise to the action following some noise schedule. Let $\ba_{t,\tau}$ denote the noisy action at diffusion step $\tau \in [0,1]$, such that:
\begin{align*}
   \ba_{t,0} = \ba_t; \qquad \ba_{t,\tau} \sim \mathcal{N}(\ba_{t}, \sigma^2_\tau I). 
\end{align*}
We choose a geometric noise schedule $\sigma_\tau = \sigma_\text{min} \left(\frac{\sigma_\text{max}}{\sigma_\text{min}}\right)^\tau$, where $\sigma_\text{min}$ and $\sigma_\text{max}$ are hyperparameters. We found it sufficient to set $\sigma_\text{min} = 10^{-5}$ and $\sigma_\text{max} = 1.0$ for all our experiments.

\paragraph{Reverse process.} Given noisy action samples, we iteratively denoise them using a learned score function to produce a sample from the target action distribution $\pi(\cdot | \bs_t) \propto \exp(Q^\pi(\bs_t, \cdot))$. 
We train a neural network, $f_\phi$ to match iDEM's $K$-sample Monte Carlo estimator of the score, defined in \Cref{eq:idem}, by setting the negative Q-function as the energy function. The score function takes as input the noisy action and diffusion time, while also being conditioned on the current state. The loss function is given by:
\begin{align}
    J(\phi) = \mathbb{E}_{\substack{(\bs_t,\ba_t) \sim \mathcal{D}, \tau \sim U[0,1],\\ \ba_{t,\tau} \sim \mathcal{N}(\ba_{t}, \sigma^2_\tau I)}}\left[\left\|f_\phi(\bs_t, \ba_{t,\tau}, \tau) - \nabla_{\ba_{t,\tau}} \log \sum_{i=1}^K \exp(Q_\theta(\bs_t, \tilde{\ba}^{(i)}_{t}))\right\|^2\right],
\label{eq:dem_loss}
\end{align}
\begin{align*}
    \ba_{t,\tau - \Delta \tau} = \ba_{t,\tau} - \sigma_\tau^2 f_\phi(\bs_t, \ba_{t,\tau}, \tau) + \sigma_\tau \epsilon,   \quad \epsilon \sim \mathcal{N}(0,I)
\end{align*}
where $\tilde{\ba}^{(i)}_{t} \sim \mathcal{N}(\ba_{t,\tau}, \sigma^2_\tau I)$.

Summarizing, to sample an action $\ba_t$ for the current state $\bs_t$ such that $\pi_\phi(\ba_t | \bs_t) \propto \exp(Q^{\pi_\phi}(\bs_t,\ba_t))$, we first sample noise from the prior (corresponding to diffusion time $\tau = 1$) $\ba_{t,1} \sim \mathcal{N(}0, \sigma^2_1)$. We then use \Cref{eq:reverse} in the VE setting (i.e. $\alpha(\tau)=0$) by using the trained score function $f_\phi$ in place of $\nabla \log p_\tau$ to iteratively denoise samples produce the action sample $\ba_t$. The full algorithm is presented in Algorithm \ref{alg:dqs}.

\paragraph{Temperature.} 
We can incorporate the temperature parameter $T$ from \Cref{eq:boltzmann} within our framework by simply scaling the Q-function in \Cref{eq:dem_loss} and regressing to the estimated score of the temperature-scaled Boltzmann distribution. To enable the score network to model this temperature scaling accurately, we additionally condition $f_\phi$ on the current temperature. Generally, the temperature is set to a high value initially, and is annealed over time such that at $t \rightarrow \infty$, we have $T \rightarrow 0$. In practice, the temperature is annealed to a sufficiently small value for large time steps. This ensures that the policy explores initially and as it collects more information about the environment, starts exploiting more and more as time passes.

\subsection{An illustrative experiment}

In this section, we aim to answer the question: \textit{does \dqs effectively learn a Boltzmann policy?}

Since the policy learning is based on iDEM, one may assume that simply optimizing \Cref{eq:dem_loss} should produce a policy that samples from the Boltzmann distribution of $Q(s,a)$. However, the learning dynamics in the interactive setting are fundamentally different from the sampling setting.

In the sampling case, the diffusion model is trained to sample from a \textit{known, fixed} energy function. This means that the target score in \Cref{eq:dem_loss} corresponds to a static function. In the actor-critic algorithm described above, the diffusion policy tracks the Q-function of the current policy. Since this Q-function is learned simultaneously along with the policy, it is a \textit{moving} target, hence it is not immediately obvious whether the proposed algorithm leads to stable learning.

We test \dqs in a controlled setting where we can qualitatively and quantitatively compare with a known ground truth distribution. Consider 2-dimensional state space $(x,y)$, an action space $(\Delta x, \Delta y)$ with the actions normalized to be unit length, and an episode length of $100$ steps. The reward function is the log likelihood of samples under a mixture of 40 Gaussian distributions. We train a Q-function and diffusion policy using \dqs (Algorithm \ref{alg:dqs}) and measure the Maximum Mean Discrepancy (MMD) between the final policy samples and the ground truth samples. \Cref{fig:gmm} plots the final samples at the end of the episode and the log partition function $\log Z = \log \int_\ba Q(\bs_t,\ba) \mathrm{d}\ba$. To demonstrate the benefit of using a diffusion-based policy, we use soft actor-critic (SAC) with a Gaussian policy as a representative baseline. We observe that \dqs approximates the ground truth samples more closely and covers most modes, which is also corroborated by a lower MMD compared to SAC.

\section{Experiments}

We perform experiments to answer the following major questions:
\begin{itemize}
    \item Does \dqs offer improved sample efficiency in continuous control tasks?
    \item Can \dqs learn multimodal behaviors, i.e., learn multiple ways to solve a task?
\end{itemize}

\paragraph{Baselines.} We test our method against a number of relevant methods. This includes classical RL algorithms such as (1) Soft Actor-Critic (SAC) \citep{haarnoja2018softa}, a maximum entropy RL method that is widely used for continuous state-action spaces; (2) Deep Deterministic Policy Gradients (DDPG) \citep{Lillicrap2015Continuous} which uses a deterministic policy and directly backpropagates gradients through the Q-function; (3) Proximal Policy Optimization (PPO) \citep{schulman2017proximal}, an on-policy policy gradient algorithm that uses a clipped objective for stable updates; and (4) Twin Delayed DDPG (TD3) \citep{fujimoto2018addressing}, an off-policy actor-critic method that mitigates overestimation bias by training two critics and delaying policy updates.

We also compare against some recent diffusion-based RL algorithms, including 
(5) Q-Score Matching (QSM) \citep{psenka2023learning}, a diffusion-based approach that trains a score function to match the gradient of the Q-function and uses this score function to sample actions; (6) Diffusion Actor-Critic with Entropy Regulator (DACER) \citep{wang2024diffusion}, which uses a diffusion-based maximum entropy policy along with Gaussian mixture models to estimate entropy; and (7) Diffusion Policy (DIPO) \citep{yang2023policy} which samples actions from a diffusion policy and performs gradient ascent using Q-functions to improve the actions.

All methods were trained with $250k$ environment interactions and one network update per environment step. For a fair comparison, all policy/score networks are MLPs with two hidden layers of dimension $256$ each, and the learning rate for all networks is $3\times 10^{-4}$. We tune hyperparameters for each baseline and select the one that gives the overall best performance across tasks. We apply the double Q-learning trick, a commonly used technique, where two Q-networks are trained independently and their minimum value is used for policy evaluation to avoid overestimation bias.

\subsection{Continuous control tasks}

\begin{figure}[!h]
    \centering
    \includegraphics[width=\linewidth]{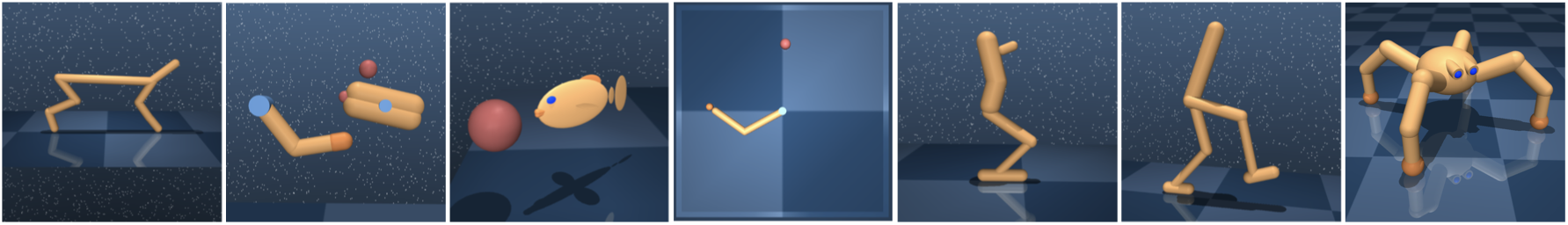}
    \caption{Domains from DeepMind Control Suite considered in our experiments - cheetah, finger, fish, reacher, hopper, quadruped and walker.}
    \label{fig:dmc_tasks}
\end{figure}

We evaluate the performance of \dqs on several continuous control tasks via the DeepMind Control Suite. We choose eight tasks from different domains to cover tasks of varying complexity and dynamics. These tasks typically involve controlling the torques applied at the joints of robots to reach a specific configuration or location, or for locomotion.

\begin{figure}[!t]
    \centering
    \includegraphics[width=\linewidth]{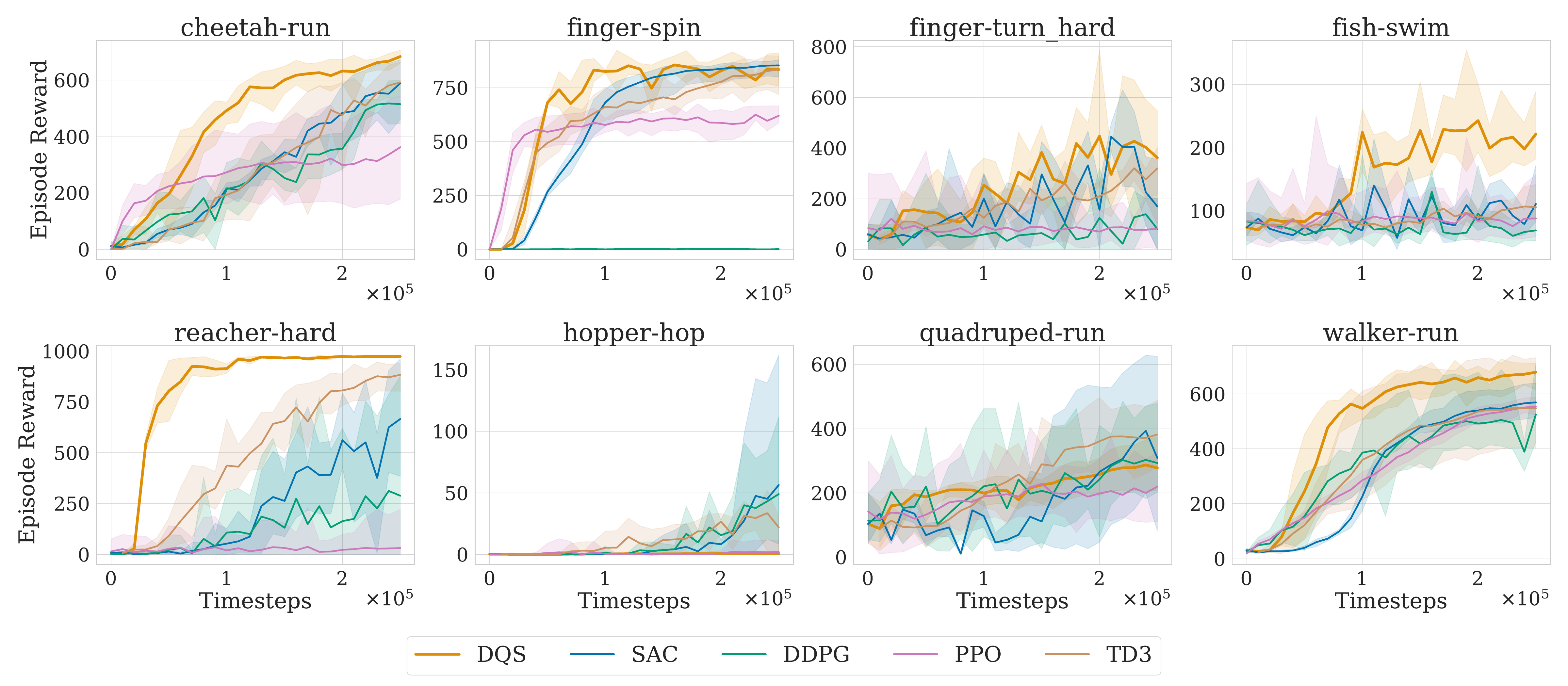}
    \caption{Experimental results for classic RL algorithms on 8 tasks from different domains from the DeepMind Control Suite. Each result is averaged over 100 evaluation episodes across 10 seeds, with the shaded regions showing minimum and maximum values. For PPO, the x-axis represents the number of network updates.}
    \label{fig:dmc_classic}
\end{figure}

\begingroup
\setlength{\tabcolsep}{4pt}       
\renewcommand{\arraystretch}{1.05} 
\begin{table}[!t]
\caption{Mean episode returns for 100 evaluation episodes for classic RL algorithms, averaged over 10 seeds. The highest mean values in each row are \colorbox{blue!10}{highlighted} and values within one standard deviation are \underline{underlined}.}
\label{table:dmc_classic}
\begin{center}
\begin{adjustbox}{width=\textwidth}
\begin{tabular}{l l l l l l l}
\toprule
 & \multicolumn{1}{c}{\textbf{Task}} & \multicolumn{1}{c}{\textbf{DQS (Ours)}} & \multicolumn{1}{c}{\textbf{SAC}} & \multicolumn{1}{c}{\textbf{DDPG}} & \multicolumn{1}{c}{\textbf{PPO}} & \multicolumn{1}{c}{\textbf{TD3}} \\ \midrule
\multirow{8}{*}{\rotatebox[origin=c]{90}{100k steps}}\hspace{1em} & cheetah-run & \highlight{492.61\scriptsize{$\pm  30.99$}} & 216.88 \scriptsize{$\pm  18.51$} & 214.02 \scriptsize{$\pm  64.55$} & 273.52 \scriptsize{$\pm  101.92$} & 193.63 \scriptsize{$\pm  68.43$} \\
 & finger-spin & \highlight{826.00 \scriptsize{$\pm  6.24$}} & 681.72 \scriptsize{$\pm  68.96$} & 0.36 \scriptsize{$\pm  0.64$} & 577.47 \scriptsize{$\pm  25.96$} & 661.70 \scriptsize{$\pm  61.30$} \\
 & finger-turn\_hard & \highlight{253.72 \scriptsize{$\pm  104.84$}} & \underline{200.02 \scriptsize{$\pm  85.60$}} & 58.58 \scriptsize{$\pm  119.76$} & 91.15 \scriptsize{$\pm  120.00$} & 124.59 \scriptsize{$\pm  32.54$} \\
 & fish-swim & \highlight{224.27} \scriptsize{$\pm  38.96$} & 69.31 \scriptsize{$\pm  11.95$} & 87.40 \scriptsize{$\pm  53.15$} & 83.86 \scriptsize{$\pm  29.02$} & 77.88 \scriptsize{$\pm  7.69$} \\
 & hopper-hop & 0.33 \scriptsize{$\pm  0.35$} & 0.01 \scriptsize{$\pm  0.02$} & 1.36 \scriptsize{$\pm  2.62$} & 0.50 \scriptsize{$\pm  0.99$} & \highlight{5.34} \scriptsize{$\pm  7.47$} \\
 & quadruped-run & \underline{199.26 \scriptsize{$\pm  32.79$}} & 127.94 \scriptsize{$\pm  32.06$} & \highlight{220.66} \scriptsize{$\pm  159.40$} & \underline{188.86 \scriptsize{$\pm  85.45$}} & \underline{191.11 \scriptsize{$\pm  21.99$}} \\
 & reacher-hard & \highlight{914.15} \scriptsize{$\pm  18.28$} & 52.76 \scriptsize{$\pm  50.88$} & 107.20 \scriptsize{$\pm  124.32$} & 19.63 \scriptsize{$\pm  46.40$} & 437.10 \scriptsize{$\pm  138.98$} \\
 & walker-run & \highlight{547.39} \scriptsize{$\pm  32.58$} & 226.52 \scriptsize{$\pm  37.38$} & 386.05 \scriptsize{$\pm  98.73$} & 285.89 \scriptsize{$\pm  25.10$} & 359.85 \scriptsize{$\pm  158.52$} \\
\hline
\multirow{8}{*}{\rotatebox[origin=c]{90}{250k steps}}  & cheetah-run & \highlight{683.64} \scriptsize{$\pm  18.51$} & 588.91 \scriptsize{$\pm  84.52$} & 514.91 \scriptsize{$\pm 61.35$} & 362.27 \scriptsize{$\pm  145.64$} & 592.13 \scriptsize{$\pm  49.08$} \\
 & finger-spin & \underline{835.00 \scriptsize{$\pm  61.36$}} & \highlight{854.04} \scriptsize{$\pm  31.92$} & 1.10 \scriptsize{$\pm  1.12$} & 620.32 \scriptsize{$\pm  31.55$} & \underline{835.45 \scriptsize{$\pm  71.52$}} \\
 & finger-turn\_hard & \highlight{361.46} \scriptsize{$\pm  179.32$} & 169.58 \scriptsize{$\pm  141.64$} & 80.48 \scriptsize{$\pm  80.64$} & 82.34 \scriptsize{$\pm  78.69$} & \underline{319.80 \scriptsize{$\pm  75.90$}} \\
 & fish-swim & \highlight{221.67} \scriptsize{$\pm  42.59$} & 111.09 \scriptsize{$\pm  35.72$} & 69.33 \scriptsize{$\pm  21.20$} & 88.13 \scriptsize{$\pm  39.20$} & 106.06 \scriptsize{$\pm  21.31$} \\
 & hopper-hop & 0.66 \scriptsize{$\pm  0.08$} & \highlight{56.47} \scriptsize{$\pm  61.47$} & \underline{49.10 \scriptsize{$\pm  44.49$}} & 2.00 \scriptsize{$\pm  4.71$} & 21.91 \scriptsize{$\pm  12.10$} \\
 & quadruped-run & \underline{277.63 \scriptsize{$\pm  66.45$}} & \underline{308.61 \scriptsize{$\pm  216.94$}} & \underline{292.94 \scriptsize{$\pm  110.62$}} & \underline{219.69 \scriptsize{$\pm  86.76$}} & \highlight{381.84} \scriptsize{$\pm  82.94$} \\
 & reacher-hard & \highlight{974.05} \scriptsize{$\pm  1.64$} & 666.12 \scriptsize{$\pm  230.56$} & 288.54 \scriptsize{$\pm  350.28$} & 31.21 \scriptsize{$\pm  88.40$} & 883.55 \scriptsize{$\pm  54.84$} \\
 & walker-run & \highlight{679.17} \scriptsize{$\pm  38.63$} & 569.28 \scriptsize{$\pm  50.41$} & 525.64 \scriptsize{$\pm  113.60$} & 553.79 \scriptsize{$\pm  26.28$} & 548.09 \scriptsize{$\pm  131.90$} \\
\bottomrule
\end{tabular}
\end{adjustbox}
\end{center}
\end{table}

\begin{figure}[!t]
    \centering
    \includegraphics[width=\linewidth]{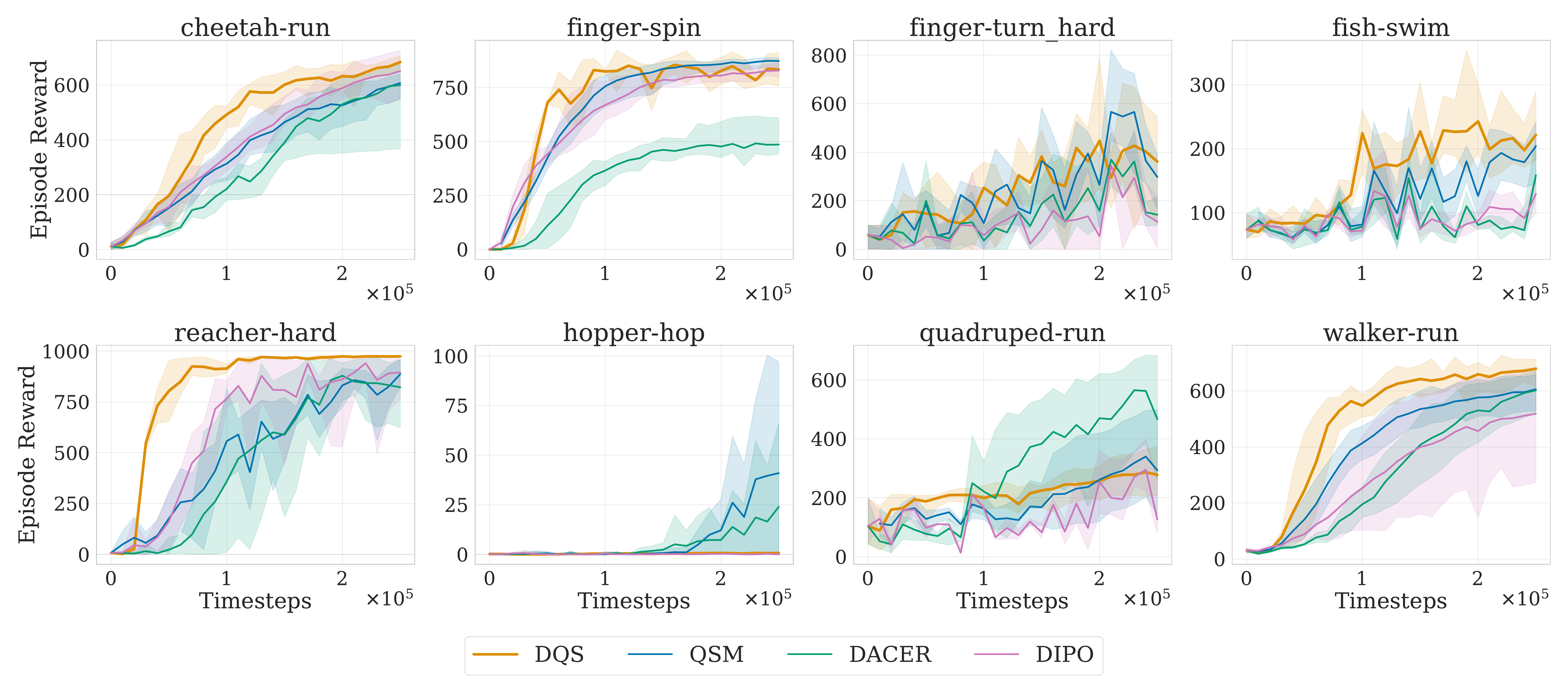}
    \caption{Experimental results for diffusion-based RL algorithms on 8 tasks from different domains from the DeepMind Control Suite. Each result is averaged over 100 evaluation episodes across 10 seeds, with the shaded regions showing minimum and maximum values.}
    \label{fig:dmc}
\end{figure}

\begingroup
\setlength{\tabcolsep}{4pt}       
\renewcommand{\arraystretch}{1.05} 
\begin{table}[!t]
\caption{Mean episode returns for 100 evaluation episodes for diffusion-based RL methods, averaged over 10 seeds. The highest mean values in each row are \colorbox{blue!10}{highlighted} and values within one standard deviation are \underline{underlined}.}
\label{table:dmc}
\begin{center}
\begin{adjustbox}{width=0.8\textwidth}
\begin{tabular}{llllll}
\toprule
 & \multicolumn{1}{c}{\textbf{Task}} & \multicolumn{1}{c}{\textbf{DQS (Ours)}} & \multicolumn{1}{c}{\textbf{QSM}} & \multicolumn{1}{c}{\textbf{DACER}} & \multicolumn{1}{c}{\textbf{DIPO}} \\ \midrule
\multirow{8}{*}{\rotatebox[origin=c]{90}{100k steps}} \hspace{1em} & cheetah-run & \highlight{492.61 \scriptsize{$\pm 30.99$}} & 313.34 \scriptsize{$\pm 55.38$} & 221.14 \scriptsize{$\pm 33.26$} & 338.93 \scriptsize{$\pm 37.90$} \\
 & finger-spin & \highlight{826.00 \scriptsize{$\pm 6.24$}} & 757.15 \scriptsize{$\pm 59.20$} & 366.06 \scriptsize{$\pm 44.68$} & 670.22 \scriptsize{$\pm 71.68$} \\
 & \text{finger-turn\_hard} & \highlight{253.72 \scriptsize{$\pm 104.84$}} & \underline{108.12 \scriptsize{$\pm 100.52$}} & 36.04 \scriptsize{$\pm 37.44$} & 57.84 \scriptsize{$\pm 48.80$} \\
 & fish-swim & \highlight{224.27 \scriptsize{$\pm 38.96$}} & 81.73 \scriptsize{$\pm 18.83$} & 78.20 \scriptsize{$\pm 14.72$} & 72.12 \scriptsize{$\pm 3.86$} \\
 & hopper-hop & \underline{0.33 \scriptsize{$\pm 0.35$}} & 0.04 \scriptsize{$\pm 0.01$} & \highlight{0.48 \scriptsize{$\pm 0.52$}} & \underline{0.41 \scriptsize{$\pm 0.59$}} \\
 & quadruped-run & \underline{199.26 \scriptsize{$\pm 32.79$}} & \underline{163.60 \scriptsize{$\pm 32.42$}} & \highlight{221.39 \scriptsize{$\pm 72.44$}} & \underline{164.08 \scriptsize{$\pm 5.37$}} \\
 & reacher-hard & \highlight{914.15 \scriptsize{$\pm 18.28$}} & 557.12 \scriptsize{$\pm 178.20$} & 358.48 \scriptsize{$\pm 303.20$} & 766.20 \scriptsize{$\pm 53.16$} \\
 & walker-run & \highlight{547.39 \scriptsize{$\pm 32.58$}} & 413.53 \scriptsize{$\pm 47.69$} & 195.30 \scriptsize{$\pm 43.95$} & 252.88 \scriptsize{$\pm 134.23$} \\
\hline
\multirow{8}{*}{\rotatebox[origin=c]{90}{250k steps}}\hspace{1em}  & cheetah-run & \highlight{683.64 \scriptsize{$\pm 18.51$}} & 607.16 \scriptsize{$\pm 36.45$} & 599.64 \scriptsize{$\pm 123.50$} & \underline{650.98 \scriptsize{$\pm 71.03$}} \\
 & finger-spin & \underline{835.00 \scriptsize{$\pm 61.36$}} & \highlight{874.52 \scriptsize{$\pm 19.38$}} & 486.66 \scriptsize{$\pm 67.52$} & \underline{830.36 \scriptsize{$\pm 27.16$}} \\
 & \text{finger-turn\_hard} & \highlight{361.46 \scriptsize{$\pm 179.32$}} & \underline{297.66 \scriptsize{$\pm 90.00$}} & 142.66 \scriptsize{$\pm 47.16$} & 113.32 \scriptsize{$\pm 85.20$} \\
 & fish-swim & \highlight{221.67 \scriptsize{$\pm 42.59$}} & \underline{204.20 \scriptsize{$\pm 38.65$}} & 158.82 \scriptsize{$\pm 37.19$} & 129.76 \scriptsize{$\pm 23.50$} \\
 & hopper-hop & 0.66 \scriptsize{$\pm 0.08$} & \highlight{40.97 \scriptsize{$\pm 38.86$}} & \underline{24.01 \scriptsize{$\pm 26.32$}} & 0.00 \scriptsize{$\pm 0.01$} \\
 & quadruped-run & \underline{277.63 \scriptsize{$\pm 66.45$}} & \underline{293.12 \scriptsize{$\pm 141.69$}} & \highlight{466.77} \scriptsize{$\pm 169.98$} & 126.09 \scriptsize{$\pm 75.00$} \\
 & reacher-hard & \highlight{974.05 \scriptsize{$\pm 1.64$}} & 887.26 \scriptsize{$\pm 54.04$} & \underline{821.88 \scriptsize{$\pm 134.16$}} & 894.88 \scriptsize{$\pm 79.32$} \\
 & walker-run & \highlight{679.17 \scriptsize{$\pm 38.63$}} & 605.98 \scriptsize{$\pm 52.12$} & 602.12 \scriptsize{$\pm 61.36$} & 518.49 \scriptsize{$\pm 162.78$} \\
\bottomrule
\end{tabular}
\end{adjustbox}
\end{center}
\end{table}

Since we are interested in evaluating the data efficiency of \dqs, we limit the number of environment interactions to $250k$ steps. \Cref{fig:dmc_classic} shows the performance of various classic methods on these different tasks. On most tasks, \dqs performs on par or outperforms the baseline methods. In particular, on five out of the eight tasks considered (cheetah-run, finger-spin, fish-swim, reacher-hard and walker-run) \dqs reaches higher reward much faster than competing methods, demonstrating improved exploration. From the numerical results in \Cref{table:dmc_classic}, we see that \dqs particularly shines in very low environment interaction budgets. When all agents are limited to $100k$ environment steps, \dqs is much more performant than other methods. Note that for PPO, the step number represents the number of network updates.

We perform a similar analysis in \Cref{fig:dmc} and \Cref{table:dmc}, where we compare the performance of \dqs with more recent diffusion-based RL methods. We describe these approaches in more detail in \Cref{sec:related}. We observe a similar trend as the previous set of experiments, where QSM achieves higher returns quicker than other methods on a majority of tasks, which is especially marked when considering performance at $100k$ environment steps. This improved sample efficiency could possibly a result of better ability to handle exploration and exploitation, owing to the use of a Boltzmann policy that samples high Q-value actions while maintaining some probability of sampling exploratory actions.

We use a single fixed temperature of $T = 0.05$ across tasks. Note that SAC and DACER use automatic temperature tuning which allows them to influence the policy entropy over the course of training. The performance of \dqs may be further improved by fine-tuning the temperature schedule on each individual task.

\subsection{Goal reaching maze navigation}

\begin{figure*}[!t]
  \centering
  \begin{subfigure}[b]{0.19\linewidth}
      \centering
      \includegraphics[width=\linewidth]{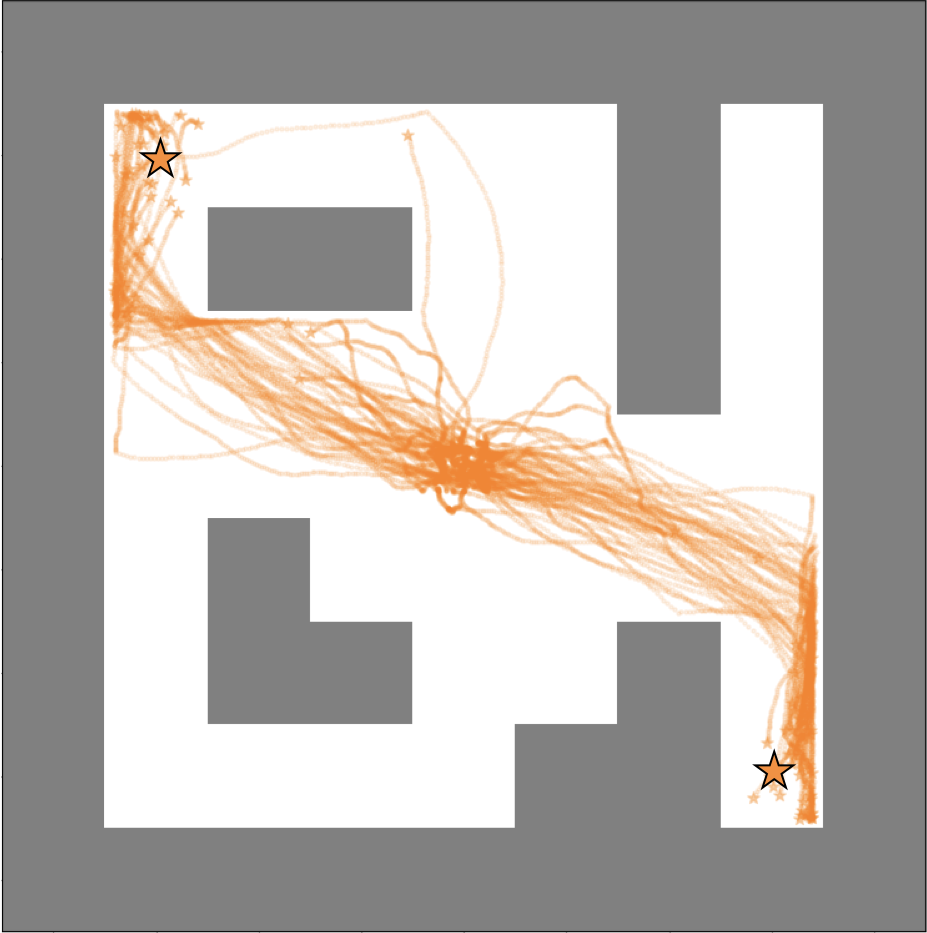}
      \caption{\textbf{DQS (ours)}}
      \label{fig:dqs_maze}
  \end{subfigure}
  \begin{subfigure}[b]{0.19\linewidth}
      \centering
      \includegraphics[width=\linewidth]{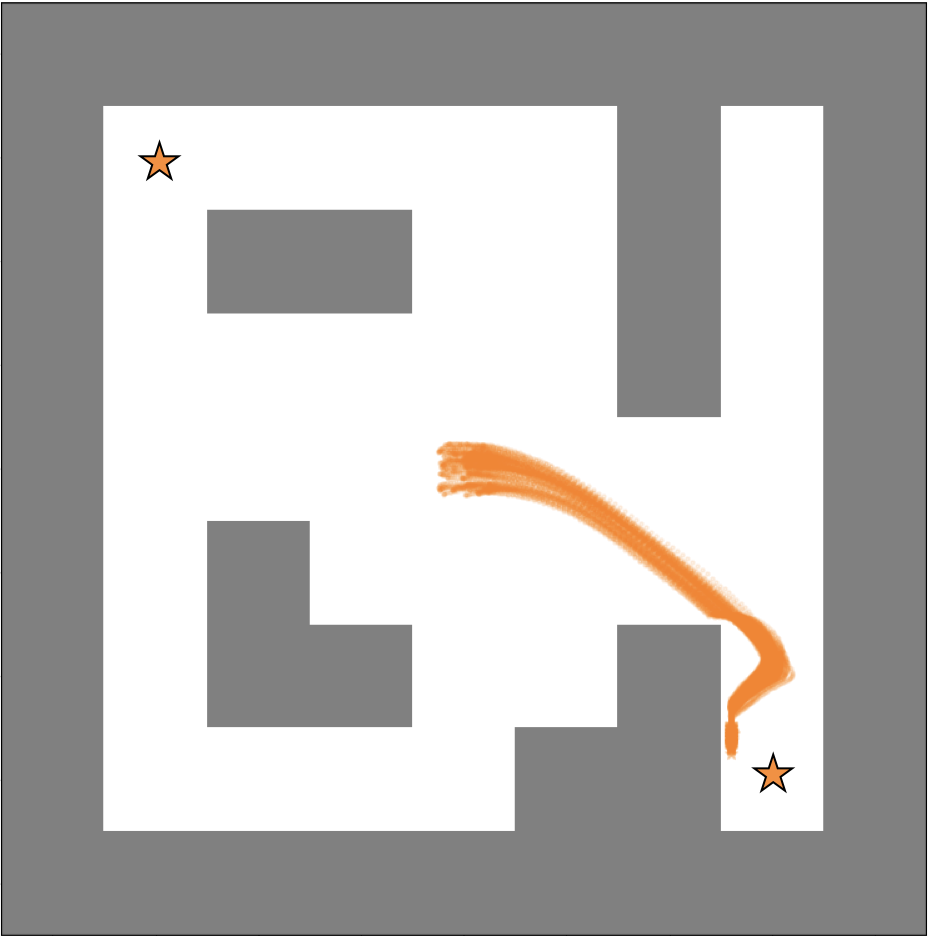}
      \caption{SAC}
      \label{fig:sac_maze}
  \end{subfigure}
  \begin{subfigure}[b]{0.19\linewidth}
      \centering
      \includegraphics[width=\linewidth]{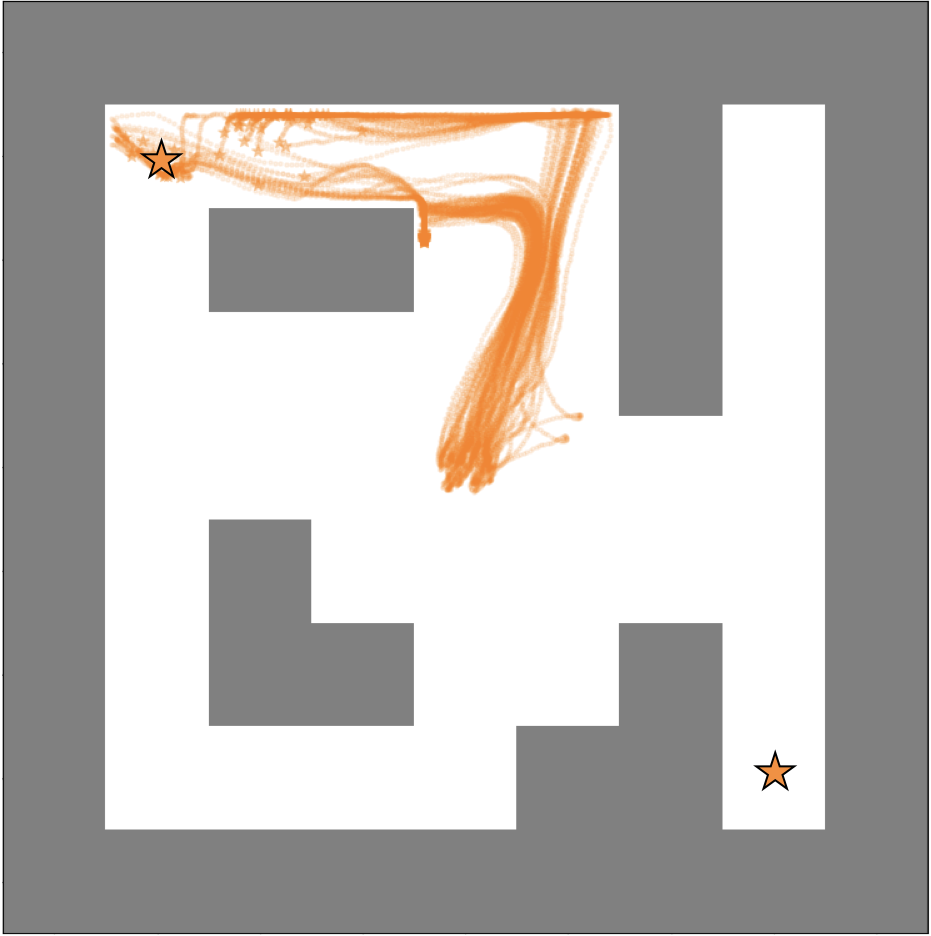}
      \caption{QSM}
      \label{fig:qsm_maze}
  \end{subfigure}
  \centering
  \begin{subfigure}[b]{0.19\linewidth}
      \centering
      \includegraphics[width=\linewidth]{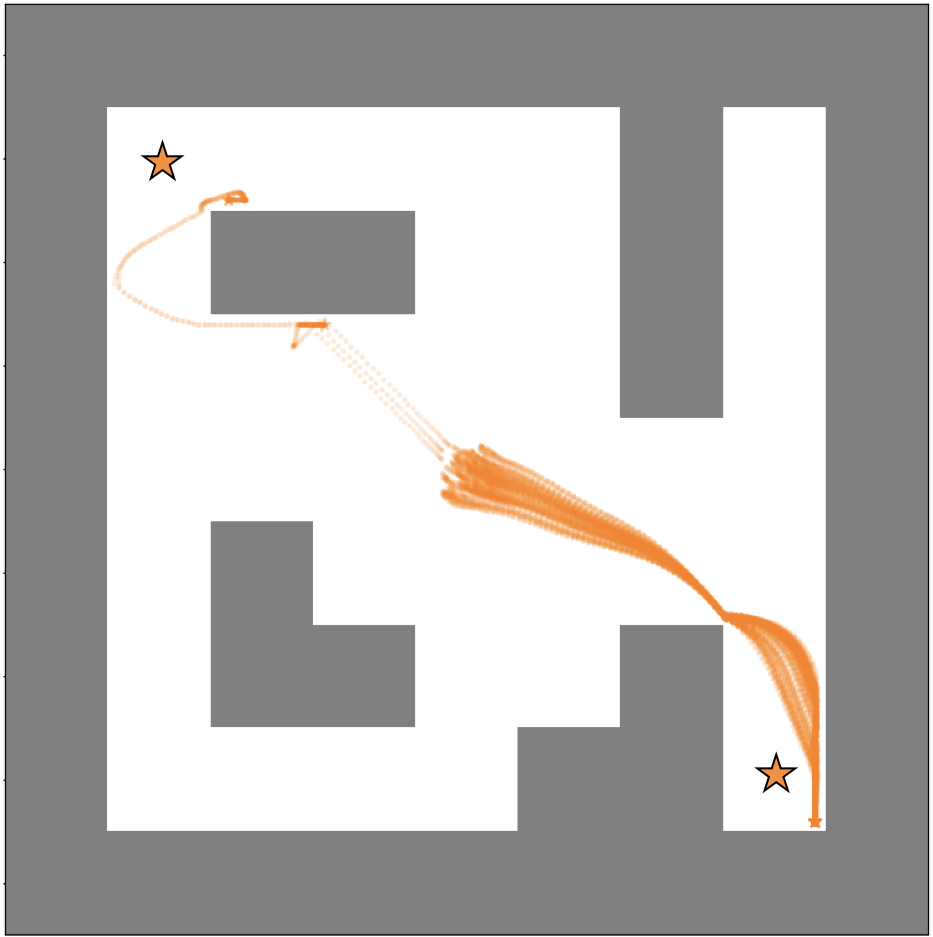}
      \caption{DACER}
      \label{fig:dacer_maze}
  \end{subfigure}
  \begin{subfigure}[b]{0.19\linewidth}
      \centering
      \includegraphics[width=\linewidth]{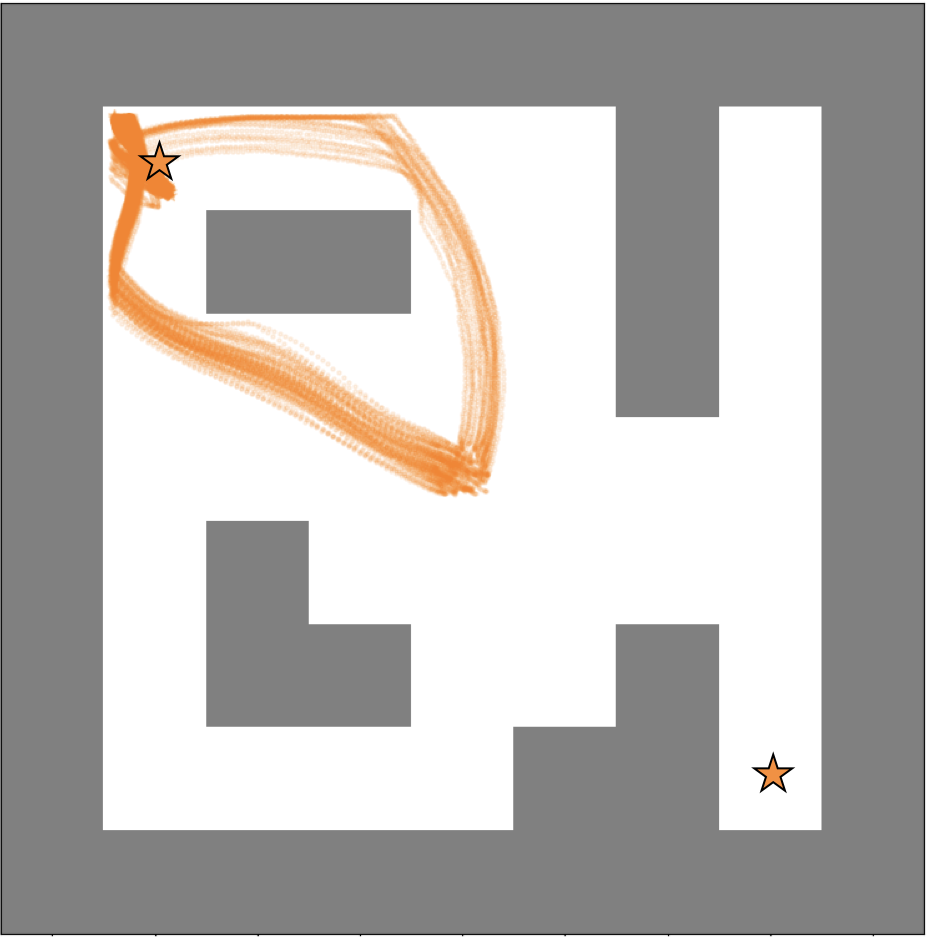}
      \caption{DIPO}
      \label{fig:dipo_maze}
  \end{subfigure}
  \caption{Trajectories for 100 evaluation episodes after $250k$ training steps. The starting states are sampled from a Gaussian distribution centered at $(0,0)$}
  \label{fig:maze}
\end{figure*}

We use a custom maze environment to evaluate the ability of our method to reach multiple goals in a sparse reward setting. The agent is tasked with manipulating a ball to reach some unknown goal position in the maze. The state consists of the ball's $(x,y)$ position and the velocity vector. The action is the force vector applied to the ball. 

The initial state of the ball is at the center of the maze, with some noise added for variability. We define two potential goal states for the ball - the top left and the bottom right corners, respectively. The reward is binary: $1.0$ is the agent reaches the goal within a radius of $0.5$m, and zero everywhere else.

For \dqs, we used temperature annealing with an initial temperature of $T = 10$, which is decayed exponentially with the number of training steps to a value of $T = 1$ after $250k$ steps. SAC uses automatic temperature tuning \citep{haarnoja2018softb}, where the entropy coefficient is automatically tuned using gradient descent to maintain the desired level of entropy.

\Cref{fig:maze} plots the trajectories of the ball over $100$ evaluation episodes after $250k$ training steps. As seen in \Cref{fig:dqs_maze}, \dqs learns to reach both goals, owing to the proposed sampling approach, which can effectively capture multimodal behavior. Moreover, it discovers both paths to reach the top left goal. In contrast, SAC (\Cref{fig:sac_maze}), QSM (\Cref{fig:qsm_maze}), and DIPO (\Cref{fig:dipo_maze}) can only reach one of the goals. Since SAC models the policy using a Gaussian, there is little variability between different trajectories. QSM produces slightly more varied behavior, since it uses Langevin sampling to sample actions, but ultimately fails to learn distinct behaviors. DIPO, on the other hand, manages to learn distinct paths to reach the same goal. DACER in \Cref{fig:dacer_maze} discovers the second mode but is heavily skewed towards one mode.

\section{Discussion}

In this work, we showcase the benefits of using energy-based policies as an expressive class of policies for deep reinforcement learning. Such policies arise in different RL frameworks, but their application has been limited in continuous action spaces owing to the difficulty of sampling in this setting. We alleviate this problem using a diffusion-based sampling algorithm,
Diffusion Q-Sampling (\dqs), that can sample multimodal behaviors and improve sample efficiency, possibly owing to better handling of the exploration-exploitation trade off. 

While diffusion methods offer high expressivity, they often come with increased computation. This is particularly true in the online RL setting, where using a diffusion policy means that each environment step requires multiple function evaluations to sample from the diffusion model. There is a growing body of work on efficient SDE samplers \citep{jolicoeur2021gotta}, which aim to reduce the number of function evaluations required to obtain diffusion-based samples while maintaining high accuracy. Incorporating such techniques with Boltzmann policies can greatly reduce the computational cost, especially in high-dimensional state-action spaces.

A crucial aspect of energy-based policies is the temperature parameter, which defines the shape of the sampling distribution. Our method enables annealing of the temperature from some starting value to lower values, as is typically done when applying Boltzmann policies in deep RL. However, this temperature schedule has to be manually tuned. \citet{haarnoja2018softb} proposes an automatic temperature tuning method for SAC, which maintains the temperature so that the entropy of the current policy is close to some target entropy. While such an approach could be applied to \dqs in principle, it is computationally expensive to compute the likelihoods of samples under a diffusion model. 

Finally, as we argued in the introduction, Boltzmann policies based on their own value function are attractive choices for pre-training of RL agents for later fine-tuning and multi-task settings. We hope to investigate this exciting potential in the future.

\bibliography{main}

\begin{thebibliography}{47}
\providecommand{\natexlab}[1]{#1}
\providecommand{\url}[1]{\texttt{#1}}
\expandafter\ifx\csname urlstyle\endcsname\relax
  \providecommand{\doi}[1]{DOI: #1}\else
  \providecommand{\doi}{DOI: \begingroup \urlstyle{rm}\Url}\fi

\bibitem[Ajay et~al.(2023)Ajay, Du, Gupta, Tenenbaum, Jaakkola, and Agrawal]{ajay2023conditional}
Anurag Ajay, Yilun Du, Abhi Gupta, Joshua~B Tenenbaum, Tommi~S Jaakkola, and Pulkit Agrawal.
\newblock Is conditional generative modeling all you need for decision making?
\newblock In \emph{The Eleventh International Conference on Learning Representations}, 2023.

\bibitem[Akhound-Sadegh et~al.(2024)Akhound-Sadegh, Rector-Brooks, Bose, Mittal, Lemos, Liu, Sendera, Ravanbakhsh, Gidel, Bengio, Malkin, and Tong]{akhoundsadegh2024iterated}
Tara Akhound-Sadegh, Jarrid Rector-Brooks, Avishek~Joey Bose, Sarthak Mittal, Pablo Lemos, Cheng-Hao Liu, Marcin Sendera, Siamak Ravanbakhsh, Gauthier Gidel, Yoshua Bengio, Nikolay Malkin, and Alexander Tong.
\newblock Iterated denoising energy matching for sampling from boltzmann densities.
\newblock \emph{arXiv}, 2024.

\bibitem[Bengio et~al.(2021)Bengio, Jain, Korablyov, Precup, and Bengio]{bengio2021flow}
Emmanuel Bengio, Moksh Jain, Maksym Korablyov, Doina Precup, and Yoshua Bengio.
\newblock Flow network based generative models for non-iterative diverse candidate generation.
\newblock \emph{Advances in Neural Information Processing Systems}, 34:\penalty0 27381--27394, 2021.

\bibitem[Bengio et~al.(2023)Bengio, Lahlou, Deleu, Hu, Tiwari, and Bengio]{bengio2023gflownet}
Yoshua Bengio, Salem Lahlou, Tristan Deleu, Edward~J Hu, Mo~Tiwari, and Emmanuel Bengio.
\newblock Gflownet foundations.
\newblock \emph{The Journal of Machine Learning Research}, 24\penalty0 (1):\penalty0 10006--10060, 2023.

\bibitem[Berner et~al.(2019)Berner, Brockman, Chan, Cheung, D{\k{e}}biak, Dennison, Farhi, Fischer, Hashme, Hesse, et~al.]{berner2019dota}
Christopher Berner, Greg Brockman, Brooke Chan, Vicki Cheung, Przemys{\l}aw D{\k{e}}biak, Christy Dennison, David Farhi, Quirin Fischer, Shariq Hashme, Chris Hesse, et~al.
\newblock Dota 2 with large scale deep reinforcement learning.
\newblock \emph{arXiv preprint arXiv:1912.06680}, 2019.

\bibitem[Chi et~al.(2023)Chi, Feng, Du, Xu, Cousineau, Burchfiel, and Song]{chi2023diffusion}
Cheng Chi, Siyuan Feng, Yilun Du, Zhenjia Xu, Eric Cousineau, Benjamin Burchfiel, and Shuran Song.
\newblock Diffusion policy: Visuomotor policy learning via action diffusion.
\newblock \emph{arXiv preprint arXiv:2303.04137}, 2023.

\bibitem[Deleu et~al.(2024)Deleu, Nouri, Malkin, Precup, and Bengio]{deleu2024discrete}
Tristan Deleu, Padideh Nouri, Nikolay Malkin, Doina Precup, and Yoshua Bengio.
\newblock Discrete probabilistic inference as control in multi-path environments.
\newblock In \emph{The 40th Conference on Uncertainty in Artificial Intelligence}, 2024.

\bibitem[Ding \& Jin(2023)Ding and Jin]{dingconsistency}
Zihan Ding and Chi Jin.
\newblock Consistency models as a rich and efficient policy class for reinforcement learning.
\newblock In \emph{The Twelfth International Conference on Learning Representations}, 2023.

\bibitem[Dinh et~al.(2016)Dinh, Sohl-Dickstein, and Bengio]{dinh2016density}
Laurent Dinh, Jascha Sohl-Dickstein, and Samy Bengio.
\newblock Density estimation using real nvp.
\newblock \emph{arXiv preprint arXiv:1605.08803}, 2016.

\bibitem[Fujimoto et~al.(2018)Fujimoto, Hoof, and Meger]{fujimoto2018addressing}
Scott Fujimoto, Herke Hoof, and David Meger.
\newblock Addressing function approximation error in actor-critic methods.
\newblock In \emph{International conference on machine learning}, pp.\  1587--1596. PMLR, 2018.

\bibitem[Haarnoja et~al.(2017)Haarnoja, Tang, Abbeel, and Levine]{haarnoja2017reinforcement}
Tuomas Haarnoja, Haoran Tang, Pieter Abbeel, and Sergey Levine.
\newblock Reinforcement learning with deep energy-based policies.
\newblock In \emph{International conference on machine learning}, pp.\  1352--1361. PMLR, 2017.

\bibitem[Haarnoja et~al.(2018{\natexlab{a}})Haarnoja, Zhou, Abbeel, and Levine]{haarnoja2018softa}
Tuomas Haarnoja, Aurick Zhou, Pieter Abbeel, and Sergey Levine.
\newblock Soft actor-critic: Off-policy maximum entropy deep reinforcement learning with a stochastic actor.
\newblock In \emph{International conference on machine learning}, pp.\  1861--1870. PMLR, 2018{\natexlab{a}}.

\bibitem[Haarnoja et~al.(2018{\natexlab{b}})Haarnoja, Zhou, Hartikainen, Tucker, Ha, Tan, Kumar, Zhu, Gupta, Abbeel, et~al.]{haarnoja2018softb}
Tuomas Haarnoja, Aurick Zhou, Kristian Hartikainen, George Tucker, Sehoon Ha, Jie Tan, Vikash Kumar, Henry Zhu, Abhishek Gupta, Pieter Abbeel, et~al.
\newblock Soft actor-critic algorithms and applications.
\newblock \emph{arXiv preprint arXiv:1812.05905}, 2018{\natexlab{b}}.

\bibitem[Hansen-Estruch et~al.(2023)Hansen-Estruch, Kostrikov, Janner, Kuba, and Levine]{hansen2023idql}
Philippe Hansen-Estruch, Ilya Kostrikov, Michael Janner, Jakub~Grudzien Kuba, and Sergey Levine.
\newblock Idql: Implicit q-learning as an actor-critic method with diffusion policies.
\newblock \emph{arXiv preprint arXiv:2304.10573}, 2023.

\bibitem[Ho et~al.(2020)Ho, Jain, and Abbeel]{ho2020denoising}
Jonathan Ho, Ajay Jain, and Pieter Abbeel.
\newblock Denoising diffusion probabilistic models.
\newblock \emph{Advances in neural information processing systems}, 33:\penalty0 6840--6851, 2020.

\bibitem[Ishfaq et~al.(2025)Ishfaq, Wang, Islam, and Precup]{ishfaqlangevin}
Haque Ishfaq, Guangyuan Wang, Sami~Nur Islam, and Doina Precup.
\newblock Langevin soft actor-critic: Efficient exploration through uncertainty-driven critic learning.
\newblock In \emph{The Thirteenth International Conference on Learning Representations}, 2025.

\bibitem[Jain \& Ravanbakhsh(2024)Jain and Ravanbakhsh]{jainlearning}
Vineet Jain and Siamak Ravanbakhsh.
\newblock Learning to reach goals via diffusion.
\newblock In \emph{Forty-first International Conference on Machine Learning}, 2024.

\bibitem[Janner et~al.(2022)Janner, Du, Tenenbaum, and Levine]{janner2022planning}
Michael Janner, Yilun Du, Joshua Tenenbaum, and Sergey Levine.
\newblock Planning with diffusion for flexible behavior synthesis.
\newblock In \emph{International Conference on Machine Learning}, pp.\  9902--9915. PMLR, 2022.

\bibitem[Jolicoeur-Martineau et~al.(2021)Jolicoeur-Martineau, Li, Pich{\'e}-Taillefer, Kachman, and Mitliagkas]{jolicoeur2021gotta}
Alexia Jolicoeur-Martineau, Ke~Li, R{\'e}mi Pich{\'e}-Taillefer, Tal Kachman, and Ioannis Mitliagkas.
\newblock Gotta go fast when generating data with score-based models.
\newblock \emph{arXiv preprint arXiv:2105.14080}, 2021.

\bibitem[Kang et~al.(2024)Kang, Ma, Du, Pang, and Yan]{kang2024efficient}
Bingyi Kang, Xiao Ma, Chao Du, Tianyu Pang, and Shuicheng Yan.
\newblock Efficient diffusion policies for offline reinforcement learning.
\newblock \emph{Advances in Neural Information Processing Systems}, 36, 2024.

\bibitem[Kingma(2014)]{kingma2014adam}
Diederik~P Kingma.
\newblock Adam: A method for stochastic optimization.
\newblock \emph{arXiv preprint arXiv:1412.6980}, 2014.

\bibitem[Kober et~al.(2013)Kober, Bagnell, and Peters]{kober2013reinforcement}
Jens Kober, J~Andrew Bagnell, and Jan Peters.
\newblock Reinforcement learning in robotics: A survey.
\newblock \emph{The International Journal of Robotics Research}, 32\penalty0 (11):\penalty0 1238--1274, 2013.

\bibitem[Kostrikov et~al.(2021)Kostrikov, Nair, and Levine]{kostrikov2021offline}
Ilya Kostrikov, Ashvin Nair, and Sergey Levine.
\newblock Offline reinforcement learning with implicit q-learning.
\newblock \emph{arXiv preprint arXiv:2110.06169}, 2021.

\bibitem[Lahlou et~al.(2023)Lahlou, Deleu, Lemos, Zhang, Volokhova, Hern{\'a}ndez-Garc{\i}a, Ezzine, Bengio, and Malkin]{lahlou2023theory}
Salem Lahlou, Tristan Deleu, Pablo Lemos, Dinghuai Zhang, Alexandra Volokhova, Alex Hern{\'a}ndez-Garc{\i}a, L{\'e}na~N{\'e}hale Ezzine, Yoshua Bengio, and Nikolay Malkin.
\newblock A theory of continuous generative flow networks.
\newblock In \emph{International Conference on Machine Learning}, pp.\  18269--18300. PMLR, 2023.

\bibitem[Li et~al.(2022)Li, Li, Kabra, Srebro, Wang, and Yang]{li2022exponential}
Gene Li, Junbo Li, Anmol Kabra, Nati Srebro, Zhaoran Wang, and Zhuoran Yang.
\newblock Exponential family model-based reinforcement learning via score matching.
\newblock \emph{Advances in Neural Information Processing Systems}, 35:\penalty0 28474--28487, 2022.

\bibitem[Lillicrap et~al.(2015{\natexlab{a}})Lillicrap, Hunt, Pritzel, Heess, Erez, Tassa, Silver, and Wierstra]{Lillicrap2015Continuous}
Timothy~P Lillicrap, Jonathan~J Hunt, Alexander Pritzel, Nicolas Heess, Tom Erez, Yuval Tassa, David Silver, and Daan Wierstra.
\newblock Continuous control with deep reinforcement learning.
\newblock \emph{arXiv preprint arXiv:1509.02971}, 2015{\natexlab{a}}.

\bibitem[Lillicrap et~al.(2015{\natexlab{b}})Lillicrap, Hunt, Pritzel, Heess, Erez, Tassa, Silver, and Wierstra]{Lillicrap2015ContinuousCW}
Timothy~P. Lillicrap, Jonathan~J. Hunt, Alexander Pritzel, Nicolas Manfred~Otto Heess, Tom Erez, Yuval Tassa, David Silver, and Daan Wierstra.
\newblock Continuous control with deep reinforcement learning.
\newblock \emph{CoRR}, abs/1509.02971, 2015{\natexlab{b}}.

\bibitem[Lu et~al.(2023)Lu, Chen, Chen, Su, Li, and Zhu]{lu2023contrastive}
Cheng Lu, Huayu Chen, Jianfei Chen, Hang Su, Chongxuan Li, and Jun Zhu.
\newblock Contrastive energy prediction for exact energy-guided diffusion sampling in offline reinforcement learning.
\newblock In \emph{International Conference on Machine Learning}, pp.\  22825--22855. PMLR, 2023.

\bibitem[Madan et~al.(2023)Madan, Rector-Brooks, Korablyov, Bengio, Jain, Nica, Bosc, Bengio, and Malkin]{madan2023learning}
Kanika Madan, Jarrid Rector-Brooks, Maksym Korablyov, Emmanuel Bengio, Moksh Jain, Andrei~Cristian Nica, Tom Bosc, Yoshua Bengio, and Nikolay Malkin.
\newblock Learning gflownets from partial episodes for improved convergence and stability.
\newblock In \emph{International Conference on Machine Learning}, pp.\  23467--23483. PMLR, 2023.

\bibitem[Malkin et~al.(2022)Malkin, Jain, Bengio, Sun, and Bengio]{malkin2022trajectory}
Nikolay Malkin, Moksh Jain, Emmanuel Bengio, Chen Sun, and Yoshua Bengio.
\newblock Trajectory balance: Improved credit assignment in gflownets.
\newblock \emph{Advances in Neural Information Processing Systems}, 35:\penalty0 5955--5967, 2022.

\bibitem[Psenka et~al.(2023)Psenka, Escontrela, Abbeel, and Ma]{psenka2023learning}
Michael Psenka, Alejandro Escontrela, Pieter Abbeel, and Yi~Ma.
\newblock Learning a diffusion model policy from rewards via q-score matching.
\newblock \emph{arXiv preprint arXiv:2312.11752}, 2023.

\bibitem[Rector-Brooks et~al.(2023)Rector-Brooks, Madan, Jain, Korablyov, Liu, Chandar, Malkin, and Bengio]{rector2023thompson}
Jarrid Rector-Brooks, Kanika Madan, Moksh Jain, Maksym Korablyov, Cheng-Hao Liu, Sarath Chandar, Nikolay Malkin, and Yoshua Bengio.
\newblock Thompson sampling for improved exploration in gflownets.
\newblock \emph{arXiv preprint arXiv:2306.17693}, 2023.

\bibitem[Reuss et~al.(2023)Reuss, Li, Jia, and Lioutikov]{reuss2023goal}
Moritz Reuss, Maximilian Li, Xiaogang Jia, and Rudolf Lioutikov.
\newblock Goal-conditioned imitation learning using score-based diffusion policies.
\newblock \emph{arXiv preprint arXiv:2304.02532}, 2023.

\bibitem[Schrittwieser et~al.(2020)Schrittwieser, Antonoglou, Hubert, Simonyan, Sifre, Schmitt, Guez, Lockhart, Hassabis, Graepel, et~al.]{schrittwieser2020mastering}
Julian Schrittwieser, Ioannis Antonoglou, Thomas Hubert, Karen Simonyan, Laurent Sifre, Simon Schmitt, Arthur Guez, Edward Lockhart, Demis Hassabis, Thore Graepel, et~al.
\newblock Mastering atari, go, chess and shogi by planning with a learned model.
\newblock \emph{Nature}, 588\penalty0 (7839):\penalty0 604--609, 2020.

\bibitem[Schulman et~al.(2015)Schulman, Levine, Abbeel, Jordan, and Moritz]{Schulman2015TrustRP}
John Schulman, Sergey Levine, P.~Abbeel, Michael~I. Jordan, and Philipp Moritz.
\newblock Trust region policy optimization.
\newblock \emph{ArXiv}, abs/1502.05477, 2015.

\bibitem[Schulman et~al.(2017)Schulman, Wolski, Dhariwal, Radford, and Klimov]{schulman2017proximal}
John Schulman, Filip Wolski, Prafulla Dhariwal, Alec Radford, and Oleg Klimov.
\newblock Proximal policy optimization algorithms.
\newblock \emph{arXiv preprint arXiv:1707.06347}, 2017.

\bibitem[Shen et~al.(2023)Shen, Bengio, Hajiramezanali, Loukas, Cho, and Biancalani]{shen2023towards}
Max~W Shen, Emmanuel Bengio, Ehsan Hajiramezanali, Andreas Loukas, Kyunghyun Cho, and Tommaso Biancalani.
\newblock Towards understanding and improving gflownet training.
\newblock In \emph{International Conference on Machine Learning}, pp.\  30956--30975. PMLR, 2023.

\bibitem[Silver et~al.(2016)Silver, Huang, Maddison, Guez, Sifre, Van Den~Driessche, Schrittwieser, Antonoglou, Panneershelvam, Lanctot, et~al.]{silver2016mastering}
David Silver, Aja Huang, Chris~J Maddison, Arthur Guez, Laurent Sifre, George Van Den~Driessche, Julian Schrittwieser, Ioannis Antonoglou, Veda Panneershelvam, Marc Lanctot, et~al.
\newblock Mastering the game of go with deep neural networks and tree search.
\newblock \emph{nature}, 529\penalty0 (7587):\penalty0 484--489, 2016.

\bibitem[Song et~al.(2021)Song, Sohl-Dickstein, Kingma, Kumar, Ermon, and Poole]{song2021score}
Yang Song, Jascha Sohl-Dickstein, Diederik~P Kingma, Abhishek Kumar, Stefano Ermon, and Ben Poole.
\newblock Score-based generative modeling through stochastic differential equations.
\newblock In \emph{International Conference on Learning Representations (ICLR)}, 2021.

\bibitem[S{\"u}nderhauf et~al.(2018)S{\"u}nderhauf, Brock, Scheirer, Hadsell, Fox, Leitner, Upcroft, Abbeel, Burgard, Milford, et~al.]{sunderhauf2018limits}
Niko S{\"u}nderhauf, Oliver Brock, Walter Scheirer, Raia Hadsell, Dieter Fox, J{\"u}rgen Leitner, Ben Upcroft, Pieter Abbeel, Wolfram Burgard, Michael Milford, et~al.
\newblock The limits and potentials of deep learning for robotics.
\newblock \emph{The International journal of robotics research}, 37\penalty0 (4-5):\penalty0 405--420, 2018.

\bibitem[Sutton \& Barto(2018)Sutton and Barto]{sutton2018reinforcement}
Richard~S Sutton and Andrew~G Barto.
\newblock \emph{Reinforcement Learning: An Introduction}.
\newblock MIT Press, 2018.

\bibitem[Tiapkin et~al.(2024)Tiapkin, Morozov, Naumov, and Vetrov]{tiapkin2024generative}
Daniil Tiapkin, Nikita Morozov, Alexey Naumov, and Dmitry~P Vetrov.
\newblock Generative flow networks as entropy-regularized rl.
\newblock In \emph{International Conference on Artificial Intelligence and Statistics}, pp.\  4213--4221. PMLR, 2024.

\bibitem[Wang et~al.(2024)Wang, Wang, Jiang, Zou, Liu, Song, Wang, Xiao, Wu, Duan, et~al.]{wang2024diffusion}
Yinuo Wang, Likun Wang, Yuxuan Jiang, Wenjun Zou, Tong Liu, Xujie Song, Wenxuan Wang, Liming Xiao, Jiang Wu, Jingliang Duan, et~al.
\newblock Diffusion actor-critic with entropy regulator.
\newblock \emph{arXiv preprint arXiv:2405.15177}, 2024.

\bibitem[Wang et~al.(2023)Wang, Hunt, and Zhou]{wang2023diffusion}
Zhendong Wang, Jonathan~J Hunt, and Mingyuan Zhou.
\newblock Diffusion policies as an expressive policy class for offline reinforcement learning.
\newblock In \emph{The Eleventh International Conference on Learning Representations}, 2023.

\bibitem[Wu et~al.(2023)Wu, Escontrela, Hafner, Abbeel, and Goldberg]{wu2023daydreamer}
Philipp Wu, Alejandro Escontrela, Danijar Hafner, Pieter Abbeel, and Ken Goldberg.
\newblock Daydreamer: World models for physical robot learning.
\newblock In \emph{Conference on robot learning}, pp.\  2226--2240. PMLR, 2023.

\bibitem[Yang et~al.(2023)Yang, Huang, Lei, Zhong, Yang, Fang, Wen, Zhou, and Lin]{yang2023policy}
Long Yang, Zhixiong Huang, Fenghao Lei, Yucun Zhong, Yiming Yang, Cong Fang, Shiting Wen, Binbin Zhou, and Zhouchen Lin.
\newblock Policy representation via diffusion probability model for reinforcement learning.
\newblock \emph{arXiv preprint arXiv:2305.13122}, 2023.

\bibitem[Ziebart et~al.(2008)Ziebart, Maas, Bagnell, Dey, et~al.]{ziebart2008maximum}
Brian~D Ziebart, Andrew~L Maas, J~Andrew Bagnell, Anind~K Dey, et~al.
\newblock Maximum entropy inverse reinforcement learning.
\newblock In \emph{Aaai}, volume~8, pp.\  1433--1438. Chicago, IL, USA, 2008.

\end{thebibliography}
\bibliographystyle{rlj}

\beginSupplementaryMaterials

\section*{Implementation details}

The score function is parameterized as an MLP with two hidden layers of $256$ units each with the ReLU activation function, except for the final layer. The MLP has skip connections as is typical for denoising score functions. The input to the policy comprises the state, noised action, the diffusion time step, and the temperature. The diffusion time step and the temperature are encoded using sinusoidal positional embeddings of $256$ dimensions. The action is sampled following \Cref{eq:reverse} and the $\mathrm{tanh(\cdot)}$ function is applied to the sampled action followed by multiplication with the maximum value of the action space to ensure the value is within the correct range. The Q-network is also an MLP with two hidden layers of $256$ units each with the ReLU activation function, except for the final layer. We use two Q-networks for the double Q-learning technique, and take the minimum of the two values.

The score function and the Q-network are trained for $250k$ environment steps with one mini-batch update per environment step. Optimization is performed using the Adam optimizer \citep{kingma2014adam} with a learning rate of $3\times 10^{-4}$ and a batch size of $256$. 

\begingroup
\setlength{\tabcolsep}{10pt}       
\renewcommand{\arraystretch}{1.05} 
\begin{table}[h]
\caption{Hyperparameters.}
\label{table:hyperparam}%
\begin{center}
\begin{tabular}{lr}
\toprule
\textbf{Parameter} & \textbf{Value} \\
\midrule
Number of hidden layers & $2$ \\
Number of hidden units per layer & $256$ \\
Sinusoidal embedding dimension & $256$ \\
Activation function & ReLU \\
Optimizer & Adam \\
Learning rate & $3 \cdot 10^{-4}$ \\
Batch size & $256$ \\
Replay buffer size & $250000$ \\
Discount factor & $0.99$ \\
Gradient updates per step & $1$ \\
Target smoothing co-efficient & $0.005$ \\
Target update period & $1$ \\
Seed training steps & $10^4$ \\
$\sigma_\text{min}$ & $0.00001$ \\
$\sigma_\text{max}$ & $1$ \\
Number of Monte Carlo samples & $1000$ \\
Number of integration steps & $1000$ \\
\bottomrule
\end{tabular}
\end{center}
\end{table}

\end{document}